\documentclass[runningheads]{llncs}

\usepackage[T1]{fontenc}
\usepackage{graphicx}
\usepackage[acronym]{glossaries}
\usepackage{colortbl}
\usepackage{nicefrac}
\usepackage{subcaption}
\usepackage{booktabs}
\usepackage{xcolor}
\usepackage[linesnumbered,ruled,vlined]{algorithm2e}
\usepackage{lipsum}
\usepackage{wrapfig}
\usepackage{fontawesome}
\usepackage{tikzit}
\usepackage{multirow}
\usepackage{url}

\DontPrintSemicolon

\definecolor{myorange}{rgb}{0.906,0.435,0.317}
\definecolor{myblue}{rgb}{0.0,0.314,0.408}

\SetKwComment{Comment}{\color{green!50!black}// }{}

\SetKwProg{Function}{function}{}{}

\usepackage{amsmath, amsfonts, dsfont}

\renewcommand{\min}[1]{\underset{#1}{\text{min}}\,}

\newcommand{\iid}[0]{\overset{iid}{\sim}}
\newcommand{\argmin}[1]{\underset{#1}{\text{argmin}}\,}

\newacronym{tep}{TEP}{Tennessee Eastman Process}
\newacronym{cdfd}{CDFD}{Cross-Domain Fault Diagnosis}
\newacronym{afd}{AFD}{Automatic Fault Diagnosis}
\newacronym{mmd}{MMD}{Maximum Mean Discrepancy}
\newacronym{bcd}{BCD}{Block Coordinate Descent}
\newacronym{irl}{IRL}{Invariant Representation Learning}
\newacronym{ot}{OT}{Optimal Transport}
\newacronym{eot}{EOT}{Empirical Optimal Transport}
\newacronym{dil}{DiL}{Dictionary Learning}
\newacronym{ml}{ML}{Machine Learning}
\newacronym{erm}{ERM}{Empirical Risk Minimization}
\newacronym{da}{DA}{Domain Adaptation}
\newacronym{tl}{TL}{Transfer Learning}
\newacronym{msda}{MSDA}{Multi-Source DA}
\newacronym{nmf}{NMF}{Nonlinear Matrix Factorization}
\newacronym{sota}{SOTA}{State-of-the-Art}
\newacronym{wdl}{WDL}{Wasserstein Dictionary Learning}
\newacronym{dadil}{DaDiL}{Dataset Dictionary Learning}
\newacronym{em}{EM}{Expectation-Maximization}
\newacronym{gmm}{GMM}{Gaussian Mixture Model}
\newacronym{mw}{MW}{Mixture Wasserstein}

\newacronym{tca}{TCA}{Transfer Component Analysis}
\newacronym{otda}{OTDA}{Optimal Transport Domain Adaptation}
\newacronym{sa}{SA}{Subspace Alignment}
\newacronym{coral}{CORAL}{Correlation Alignment}
\newacronym{jdot}{JDOT}{Joint Distribution Optimal Transport}
\newacronym{wbt}{WBT}{Wasserstein Barycenter Transport}

\newacronym{dann}{DANN}{Domain Adversarial Neural Network}
\newacronym{wdgrl}{WDGRL}{Wasserstein Distance Guided Representation Learning}
\newacronym{mcd}{MCD}{Maximum Classifier Discrepancy}
\newacronym{mdd}{MDD}{Margin Disparity Discrepancy}
\newacronym{wjdot}{WJDOT}{Weighted JDOT}
\newacronym{m3sda}{M3SDA}{Moment Matching for MSDA}

\newacronym{tsne}{t-SNE}{t-Stochastic Neighbor Embeddings}
\newacronym{nn}{NN}{Neural Net}
\newacronym{svm}{SVM}{Support Vector Machine}
\newacronym{pot}{POT}{Python Optimal Transport}

\newacronym{map}{MAP}{Maximum a Posteriori}
\newacronym{sgd}{SGD}{Stochastic Gradient Descent}

\newacronym{jcpot}{JCPOT}{Joint Class Proportion and Optimal Transport}

\newacronym{fcn}{FCN}{Fully Convolutional Network}
\newacronym{gap}{GAP}{Global Average Pooling}

\newacronym{mds}{MDS}{Multi-Dimensional Scaling}
\newacronym{fdd}{FDD}{Fault Detection and Diagnosis}
\newacronym{cce}{CCE}{Cross-entropy}

\tikzstyle{trapezium}=[fill=white, draw=black, shape=trapezium, rotate=-90, minimum height=1cm]
\tikzstyle{lossbox}=[fill={rgb,255: red,202; green,206; blue,255}, draw=black, shape=rectangle, minimum height=1.2cm, minimum width=1cm, align=center]
\tikzstyle{clfbox}=[fill=white, draw=black, shape=rectangle, minimum width=1cm, minimum height=1cm]
\tikzstyle{new style 2}=[fill=white, draw=black, shape=rectangle, align=center]
\tikzstyle{domainbox}=[fill=white, draw=black, shape=rectangle, minimum width=3cm, align=center]
\tikzstyle{longbox}=[fill=white, draw=black, shape=rectangle, minimum height=4cm, minimum width=1.2cm, align=center]
\tikzstyle{rotatednode}=[rotate=90]
\tikzstyle{circularnode}=[fill=none, draw=black, shape=circle]
\tikzstyle{blue_circle}=[fill={rgb,255: red,0; green,80; blue,104}, draw=none, shape=circle, minimum width=0.5cm]
\tikzstyle{orangecircle1}=[fill={rgb,255: red,231; green,111; blue,81}, draw=none, shape=circle, minimum width=0.5cm]
\tikzstyle{blue_square1}=[fill={rgb,255: red,0; green,80; blue,104}, draw=none, shape=rectangle, minimum width=0.5cm, minimum height=0.5cm]
\tikzstyle{blue_triangle1}=[fill={rgb,255: red,0; green,80; blue,104}, draw=none, shape=regular polygon, regular polygon sides=3]
\tikzstyle{widebox}=[fill=white, draw=black, shape=rectangle, minimum height=1.2cm, minimum width=10cm, align=center]
\tikzstyle{labeled domain}=[fill=none, draw={rgb,255: red,0; green,80; blue,104}, shape=circle, minimum width=1cm]
\tikzstyle{unlabeled domain}=[fill=none, draw={rgb,255: red,231; green,111; blue,81}, shape=circle, minimum width=1cm]
\tikzstyle{smallwidebox}=[fill=white, draw=black, shape=rectangle, minimum height=1.2cm, minimum width=5cm, align=center]

\tikzstyle{red edge}=[->, fill=none, draw={rgb,255: red,128; green,0; blue,0}]
\tikzstyle{blue edge}=[->, fill=none, draw={rgb,255: red,70; green,130; blue,180}]
\tikzstyle{green edge}=[->, fill=none, draw={rgb,255: red,44; green,160; blue,44}]
\tikzstyle{red dotted edge}=[->, dashed, fill=none, draw={rgb,255: red,128; green,0; blue,0}, thick]
\tikzstyle{blue dotted edge}=[->, dashed, fill=none, draw={rgb,255: red,70; green,130; blue,180}]
\tikzstyle{green dotted edge}=[->, dashed, fill=none, draw={rgb,255: red,44; green,160; blue,44}, thick]
\tikzstyle{red dotted line}=[-, fill=none, dashed, draw={rgb,255: red,128; green,0; blue,0}]
\tikzstyle{blue dotted line}=[-, fill=none, dashed, draw={rgb,255: red,70; green,130; blue,180}]
\tikzstyle{green dotted line}=[-, fill=none, dashed, draw={rgb,255: red,44; green,160; blue,44}]
\tikzstyle{black edge}=[->]
\tikzstyle{black dashed line}=[-, dashed]
\tikzstyle{thick black arrow}=[->, thick]
\tikzstyle{thick black edge}=[-, thick]
\tikzstyle{thick black dotted line}=[->, thick, dashed]
\tikzstyle{semi transparent dashed black line}=[-, opacity=0.2, dashed]

\begin{document}
\title{Benchmarking Domain Adaptation for Chemical Processes on the Tennessee Eastman Process}
\titlerunning{Domain Adaptation for Chemical Processes}
%
\author{Eduardo Fernandes Montesuma\inst{1} \and
Michela Mulas\inst{2} \and
Fred Ngol\`e Mboula\inst{1} \and
Francesco Corona\inst{3} \and
Antoine Souloumiac\inst{1}}
%
\authorrunning{Montesuma, Mulas, Mboula, Corona and Souloumiac}
%
\institute{Université Paris-Saclay, CEA, LIST, F-91120, Palaiseau, France \and
Department of Teleinformatics Engineering, Federal University of
Ceará, Brazil \and
School of Chemical Engineering, Aalto University, Finland}
%
\maketitle              

\begin{abstract}
In system monitoring, automatic fault diagnosis seeks to infer the systems' state based on sensor readings, e.g., through machine learning models. In this context, it is of key importance that, based on historical data, these systems are able to generalize to incoming data. In parallel, many factors may induce changes in the data probability distribution, hindering the possibility of such models to generalize. In this sense, domain adaptation is an important framework for adapting models to different probability distributions. In this paper, we propose a new benchmark, based on the Tennessee Eastman Process of Downs and Vogel (1993), for benchmarking domain adaptation methods in the context of chemical processes. Besides describing the process, and its relevance for domain adaptation, we describe a series of data processing steps for reproducing our benchmark. We then test 11 domain adaptation strategies on this novel benchmark, showing that optimal transport-based techniques outperform other strategies.\footnote{ {\faGithub} Our code is open sourced at \url{https://github.com/eddardd/tep-domain-adaptation}}.
\end{abstract}
\section{Introduction}

Within process supervision, faults are unpermitted deviations of a characteristic property or variables of a system~\cite{isermann2006fault}. Furthermore, there is an increasing demand on reliability and safety of technical plants, motivating the necessity of methods for supervision and monitoring. These are \gls{fdd} methods, which comprise the \emph{detection}, i.e., if and when a fault has occurred, and the \emph{diagnosis}, i.e., the determination of \emph{which fault} has occurred. In this paper, we focus on \gls{afd} systems, assuming that faults were previously detected accordingly.

In parallel, \gls{ml} is a field of artificial intelligence, that defines predictive models based on data. Nonetheless, these models make an implicit assumption, that training and test data come from the same probability distribution, which is seldom verified in practice~\cite{quinonero2009dataset}, as both training and test data may be collected under heterogeneous conditions that drive shifts in probability distributions. This phenomenon motivates the field of \gls{tl}~\cite{pan2010domain} to propose algorithms that are robust to \emph{distributional shift}.

There is a straightforward link between \gls{ml} and \gls{afd} systems, as one can understand fault diagnosis as a classification problem. In this sense, one uses sensor data (e.g., temperature, concentration, flow-rate) as inputs to a classifier, which predicts the corresponding fault, or its absence~\cite{zheng2019cross}. Further, \gls{tl} is a broad field within \gls{ml}, in which knowledge must be \emph{transferred} from a source to a target context. Within \gls{tl}, \gls{da} is a common framework where one has access to labeled data from a source domain, and unlabeled data from a target domain. Thus, \gls{da} seeks improving classification accuracy on target domain data. In many cases, source data is itself heterogeneous, following multiple probability distributions. This setting is known as \gls{msda}.

In this paper, we propose a new benchmark, based on the \gls{tep}~\cite{downs1993plant,reinartz2021extended}, a complex, large-scale chemical process used by the chemical engineering community for benchmarking control systems, as well as \gls{fdd} techniques. This process is interesting for \gls{da}, as it may operate at different modes of production. As we show in our case study (section~\ref{sec:case_study}), the different modes of production induce different data probability distributions, thus the need for \gls{da} techniques for improving generalization. We further benchmark existing techniques in \gls{da}, either based on pre-extracted features (shallow \gls{da}), or through deep learning (deep \gls{da}).

The rest of this paper is divided as follows. Section~\ref{sec:preliminaries} covers the theoretical foundations of our work. Section~\ref{sec:case_study} presents a case study of the \gls{tep}. In this section, we present the system, analyze the properties of the different modes of production, and benchmark different strategies in \gls{da}. Finally, section~\ref{sec:conclusion} concludes this paper.
\section{Classification and Domain Adaptation}\label{sec:preliminaries}

In supervised learning, one is provided with a dataset $\{\mathbf{x}_{i}^{(P)},y_{i}^{(P)}\}_{i=1}^{n}$, where $\mathbf{x}_{i}^{(P)} \iid P$, and $y_{i}^{(P)} = h_{0}(\mathbf{x}_{i}^{(P)})$, for a distribution $P$ and a ground-truth labeling function $h_{0}:\mathcal{X} \rightarrow \mathcal{Y}$. $\mathcal{X}$ is called \emph{feature space}, such as $\mathbb{R}^{d}$, and $\mathcal{Y}$ \emph{label space}, in this case $\{1,\cdots,n_{c}\}$. The goal of classification is finding, among a family of functions $\mathcal{H}$, $\hat{h}$ such that,
\begin{align}
    \hat{h} = \argmin{h \in \mathcal{H}}\dfrac{1}{n}\sum_{i=1}^{n}\mathcal{L}(h(\mathbf{x}_{i}^{(P)}), y_{i}^{(P)}),\label{eq:erm}
\end{align}
where $\mathcal{L}$ is a loss function, such as the \gls{cce}, $\text{CCE}(\mathbf{y},\hat{\mathbf{y}}) = \sum_{c=1}^{n_{c}}y_{c}\log\hat{y}_{c}$. This approach, known as empirical risk minimization, has the desirable property that $\hat{h}$ correctly predicts on unseen samples from $P$. This property is known as \emph{generalization}. We refer readers to~\cite{redko2019advances} for a review on the theory of generalization.

In this paper, we consider deep neural nets composed of 2 parts: an encoder network $\phi$, and a classifier $h$. The encoder maps data $\mathbf{x} \in \mathcal{X}$ into a latent representation $\mathbf{z} \in \mathcal{Z}$, whereas the classifier maps the representation into a label space $\mathcal{Y}$. Hence, $\hat{y}_{i}^{(P)} = h(\phi(\mathbf{x}_{i}^{(P)}))$. As such, eq.~\ref{eq:erm} is minimized with respect the parameters of the encoder, $\theta_{\phi}$ and classifier $\theta_{h}$.

\begin{figure}[ht]
    \centering
    \includegraphics[width=0.7\linewidth]{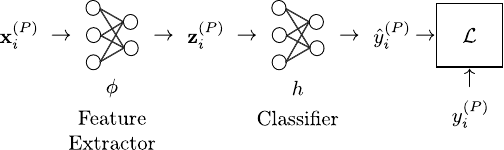}
    \caption{\textbf{Illustration of a deep neural net,} where data $\mathbf{x}_{i}^{(P)}$ are mapped into latent representation vectors $\mathbf{z}_{i}^{(P)}$ through an encoder $\phi$. The latent representation is then used to predict a class, i.e., $\hat{y}_{i}^{(P)}$.}
    \label{fig:deep_nn}
\end{figure}

The main challenge faced by \gls{ml} models is generalizing beyond samples from $P$. In this sense, it is desirable that $\hat{h}$ generalizes to different, but related distributions $Q$, which is known as \gls{tl}~\cite{pan2010domain}. Within \gls{tl}, \gls{da} is a popular framework where one seeks to improve performance on a target domain based on knowledge available in a source domain. Especially, a domain is a pair $\mathcal{D}=(\mathcal{X},P)$, where $P$ is a distribution the feature space $\mathcal{X}$. Likewise, a task is a pair $\mathcal{T}=(\mathcal{Y},h_{0})$, where $h_{0}:\mathcal{X}\rightarrow\mathcal{Y}$ is a ground-truth labeling function. Given a source domain and task $(\mathcal{D}_{S},\mathcal{T}_{S})$, and a target domain and task, $(\mathcal{D}_{T},\mathcal{T}_{T})$, in \gls{da} one has $\mathcal{T}_{S} = \mathcal{T}_{T}$, but $P_{S} \neq P_{T}$. As a consequence, $\mathcal{D}_{S} \neq \mathcal{D}_{T}$. The goal of \gls{da} can be summarized as follows: given labeled samples from the source domain, and unlabeled samples from the target domain, find a classifier $\hat{h}$ that generalizes to samples from $P_{T}$.

In addition, one may have a scenario where source domain data is heterogeneous. In this case, one assumes that this domain is composed of several distributions, i.e., $P_{S_{1}}, \cdots, P_{S_{N}}$, for $N > 1$. This case is known in the literature as \gls{msda}. Besides the challenge of having $P_{S_{\ell}} \neq P_{T}$, one has inter-domain shifts, i.e., $P_{S_{\ell}} \neq P_{S_{\ell'}}$, for $\ell \neq \ell'$.

\begin{figure}[ht]
    \centering
    \includegraphics[width=0.7\linewidth]{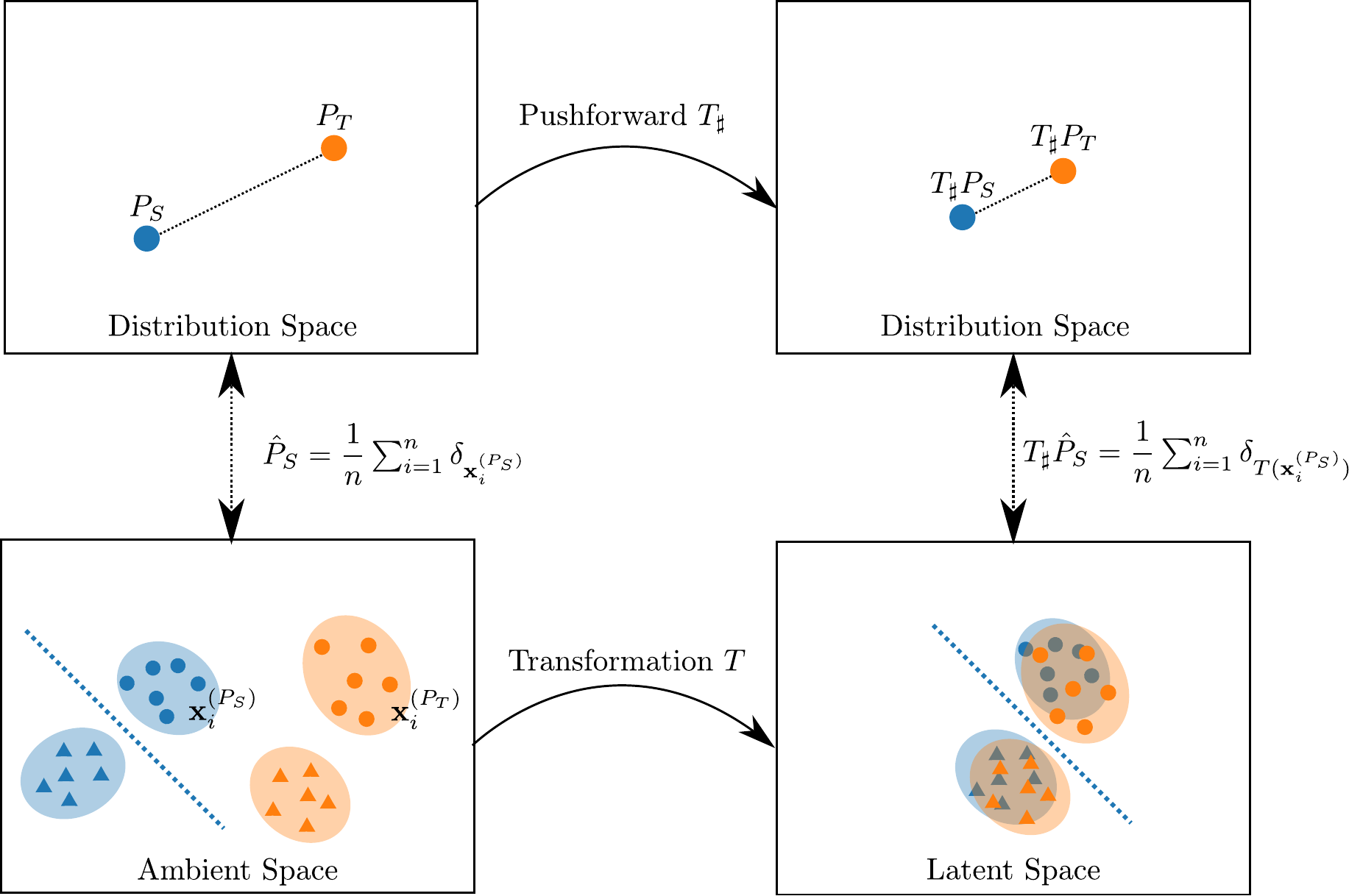}
    \caption{\textbf{Domain adaptation based on data transformation.} In an ambient space, source and target data follow different probability distributions. As a result, a classifier learned on the source (blue straight line on the left) is not able to generalize on data from the target domain (orange elements). In this paper we consider methods that align the distributions through a data transformation $T$, which maps data into a latent space.}
    \label{fig:illustration_da}
\end{figure}

Given our discussion so far, one needs a notion of \emph{closeness} between $P_{S}$ and $P_{T}$ for having generalization to new distributions $Q$~\cite[Theorem 10]{redko2019advances}. We thus focus on \gls{da} methods that seek to reduce the distance between distributions $P_{S}$ and $P_{T}$ through data transformations. In a nutshell, these methods apply a mapping to $\mathbf{x}_{i}^{(P_{S})}$, so that $\{T(\mathbf{x}_{i}^{(P_{S})})\}_{i=1}^{n}$ is distributed in the same way as $\{\mathbf{x}_{j}^{(P_{T})}\}$. This idea is illustrated in Fig.~\ref{fig:illustration_da}. This alignment supposes a criterion of \emph{dissimilarity} between these objects. In this sense, one may use \emph{probability metrics}, which are distances in the space of probability distributions. In our experiments, we consider three prominent metrics, namely, the $\mathcal{H}$-distance, the \gls{mmd} and the Wasserstein distance.

The $\mathcal{H}$-distance has its roots on \gls{da} theory~\cite{ben2010theory}. This distance measures how likely a classifier can separate samples from these distributions. Hence, let $h \in \mathcal{H}$ be a classifier,
\begin{align}
    d_{\mathcal{H}}(\hat{P}_{S},\hat{P}_{T}) = 2\biggr(1-\min{h\in\mathcal{H}}\biggr(\dfrac{1}{n}\sum_{i=1}^{n}\log(1-h(\mathbf{x}_{i}^{(P_{S})})) + \dfrac{1}{m}\sum_{j=1}^{m}\log h(\mathbf{x}_{j}^{(P_{T})})\biggr)\biggr).
\end{align}
Note that the $d_{\mathcal{H}}$ can be easily estimated from samples $\{\mathbf{x}_{i}^{(P_{S})}\}_{i=1}^{n}$ and $\{\mathbf{x}_{j}^{(P_{T})}\}_{j=1}^{m}$, by learning a classifier that predicts the domain of a given sample (e.g., $0$ for $P_{S}$, and $1$ for $P_{T}$).

The \gls{mmd} has its roots on kernel theory~\cite{gretton2007kernel}, and was initially proposed to test if two samples come from the same distribution. Let $k:\mathbb{R}^{d}\times\mathbb{R}^{d}\rightarrow\mathbb{R}$ be a kernel, the \gls{mmd} can be defined as,
\begin{equation}
\begin{aligned}
    \text{MMD}_{k}(\hat{P}_{S},\hat{P}_{T})^{2} = \dfrac{1}{n^{2}}\sum_{i=1}^{n}\sum_{j=1}^{n}k(\mathbf{x}_{i}^{(P_{S})},\mathbf{x}_{j}^{(P_{S})}) &+ \dfrac{1}{m^{2}}\sum_{i=1}^{m}\sum_{j=1}^{m}k(\mathbf{x}_{i}^{(P_{T})},\mathbf{x}_{j}^{(P_{T})})\\
    &- \dfrac{2}{nm}\sum_{i=1}^{n}\sum_{j=1}^{m}k(\mathbf{x}_{i}^{(P_{S})},\mathbf{x}_{j}^{(P_{T})}),
\end{aligned}
\end{equation}
examples of kernels include the linear kernel $k(\mathbf{x}_{i}^{(P_{S})},\mathbf{x}_{j}^{(P_{T})}) = (\mathbf{x}_{i}^{(P_{S})})^{\top}\mathbf{x}_{j}^{(P_{T})}$, and the Gaussian kernel, $k(\mathbf{x}_{i}^{(P_{S})},\mathbf{x}_{i}^{(P_{T})}) = \exp{(-\gamma\lVert \mathbf{x}_{i}^{(P_{S})} - \mathbf{x}_{j}^{(P_{T})} \rVert_{2}^{2})}$, for a parameter $\gamma > 0$. Intuitively, the \gls{mmd} is a distance between the means of distributions in an embedding space defined by the kernel $k$.

Finally, the Wasserstein distance $W_{p}$ is rooted on the theory of \gls{ot}. In its modern computational treatment \cite{peyre2019computational,montesuma2023recent}, the \gls{ot} problem can be phrased as,
\begin{equation}
\gamma^{\star} = \argmin{\gamma \in \Gamma}\sum_{i=1}^{n}\sum_{j=1}^{m}\gamma_{ij}\lVert \mathbf{x}_{i}^{(P_{S})} - \mathbf{x}_{j}^{(P_{T})}\rVert_{2}^{p},\label{eq:kantorovich}
\end{equation}
where $\gamma \in \mathbb{R}^{n\times m}$ is called \gls{ot} plan, and $\Gamma$ is the set of mass preserving plans, i.e., matrices $\gamma$ such that their row sum $\sum_{i=1}^{n}\gamma_{ij} = m^{-1}$, and column sum $\sum_{j=1}^{m}\gamma_{ij}=n^{-1}$. Problem~\ref{eq:kantorovich} is a linear program, which can be solved exactly through the Simplex method~\cite{dantzig1955generalized}. Based on $\gamma^{\star}$, the Wasserstein distance is
\begin{align*}
    W_{p}(\hat{P}_{S},\hat{P}_{T})^{p} = \sum_{i=1}^{n}\sum_{j=1}^{m}\gamma_{ij}^{\star}\lVert \mathbf{x}_{i}^{(P_{S})} - \mathbf{x}_{j}^{(P_{T})}\rVert_{2}^{p}.
\end{align*}

Let $\mathcal{N}(\mu,\Sigma)$ denote the Gaussian distribution with mean $\mu \in \mathbb{R}^{d}$, and covariance matrix $\Sigma \in \mathcal{S}^{d}_{+}$, i.e., a $d\times d$ symmetric and positive semi-definite matrix. For $p=2$,  $P_{S} = \mathcal{N}(\mu_{S},\Sigma_{S})$ and $P_{T} = \mathcal{N}(\mu_{T},\Sigma_{T})$, the Wasserstein distance is,
\begin{align}
    W_{2}(P_{S}, P_{T})^{2} = \lVert \mu_{S} - \mu_{T} \rVert_{2}^{2} + \mathcal{B}(\Sigma_{S},\Sigma_{T}),\label{eq:w2_gauss}
\end{align}
where $\mathcal{B}$ is the Bures-metric between covariance matrices~\cite{takatsu2011wasserstein}. While \gls{ot}-based \gls{da} methods use equation~\ref{eq:kantorovich}, equation~\ref{eq:w2_gauss} is commonly used for estimating the Wasserstein distance given samples. In this case, the parameters $(\mu_{S},\Sigma_{S},\mu_{T},\Sigma_{T})$ are the sample mean and covariance from each domain.

\begin{table}[ht]
\centering
\caption{Description of shallow and deep domain adaptation methods alongside the notion of distance they minimize during training.}
\begin{tabular}{cccccccc}
    \toprule
    \multicolumn{4}{c}{Single Source} & \multicolumn{4}{c}{Multi Source}\\
    Method & Distance & Category & Reference &  Method & Distance & Category & Reference\\
    \midrule
    TCA & $\text{MMD}$ & Shallow & \cite{pan2010domain} & M3SDA & $ \multirow{2}{*}{\text{MMD}} $ & \multirow{2}{*}{Deep} & \multirow{2}{*}{\cite{peng2019moment}}\\
    
    OTDA & $W_{2}$ & Shallow & \cite{courty2016otda} & M3SDA$_{\beta}$ &\\
    JDOT & $W_{2}$ & Shallow & \cite{courty2017joint} & WJDOT & $W_{2}$ & Shallow & \cite{turrisi2022multi}\\
    MMD & $\text{MMD}$ & Deep & \cite{ghifary2014domain} & WBT$_{reg}$ & $W_{2}$ & Shallow & \cite{montesuma2021icassp,montesuma2021cvpr}\\
    DANN & $d_{\mathcal{H}}$ & Deep & \cite{ganin2016domain} & DaDiL-R & \multirow{2}{*}{$W_{2}$} & \multirow{2}{*}{Shallow} & \multirow{2}{*}{\cite{montesuma2023dadil}}\\
    DeepJDOT & $W_{2}$ & Deep & \cite{damodaran2018deepjdot} & DaDiL-E & \\
    \bottomrule
\end{tabular}
\label{tab:overview_da}
\end{table}

A final distinction between \gls{da} methods is with respect their strategy. First, we consider shallow \gls{da} methods. These strategies apply transformations to pre-extracted features, in the hope of aligning the data distributions. For instance, \gls{tca}~\cite{pan2010domain} projects data into a lower dimensional space while minimizing the \gls{mmd}. With respect the architecture shown in Fig.~\ref{fig:deep_nn}, these methods keep the parameters of the encoder network $\phi$ frozen during adaptation, and fine-tune the classifier $h$ on the adapted data. Second, we consider deep \gls{da} methods, which rely on the encoder network $\phi$ for aligning the data. The principle is to minimize the distance in distribution between $\phi_{\sharp}P_{S}$ and $\phi_{\sharp}P_{T}$ (c.f., Fig.~\ref{fig:illustration_da}). This is the case of DeepJDOT~\cite{damodaran2018deepjdot}, which minimizes the Wasserstein distance between the aforementioned distributions. In total, we consider 11 methods, as shown in Table~\ref{tab:overview_da}. We refer readers to the original papers for further details on these algorithms.
\section{Case Study: the Tennessee Eastman Process}\label{sec:case_study}

\begin{figure}[ht]
    \centering
    \includegraphics[width=0.8\linewidth]{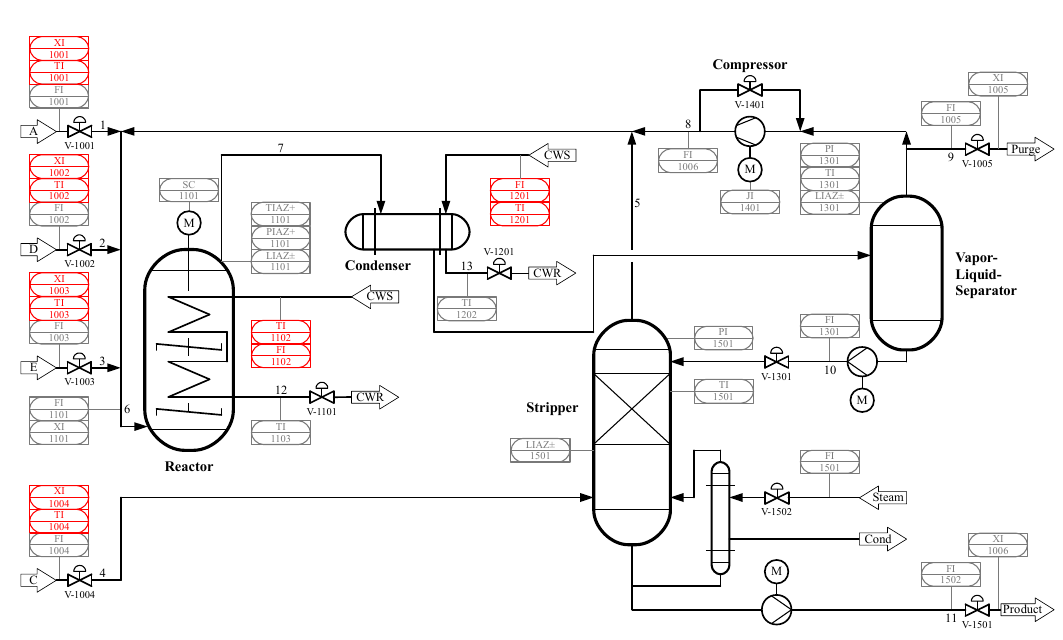}
    \caption{\textbf{P\&ID diagram for the \gls{tep}.} Figure reproduced from~\cite{bathelt2015revision}, which shows the main components of the process. Measurements originally introduced by~\cite{downs1993plant} are shown in gray, whereas the measurements introduced by~\cite{bathelt2015revision} are shown in red. A simulation environment, based on this diagram, is described in~\cite{reinartz2021extended}.}
    \label{fig:te_process_schematic}
\end{figure}

In this section, we present our case study, the Tennessee Eastman Process (TEP). This chemical process was first introduced by~\cite{downs1993plant}, with the intent to serve as a realistic benchmark for the design of control and monitoring systems. From the perspective of fault detection and diagnosis~\cite{melo2022open}, this system is widely used by the academic community. Henceforth, we follow the description of the \gls{tep} by~\cite{reinartz2021extended}. The \gls{tep} consists on the production of two liquid product components, $G$ and $H$, from $4$ gaseous reactants, $A$, $C$, $D$ and $E$, with an additional inert $B$ and a byproduct $F$, which are related through $4$ exothermic and irreversible reactions,
\begin{equation}
    \begin{aligned}
        A(\text{g}) + C(\text{g}) + D(\text{g}) &\rightarrow G(\text{liq})&\text{Product 1,}\\
        A(\text{g}) + C(\text{g}) + E(\text{g}) &\rightarrow H(\text{liq})&\text{Product 2,}\\
        A(\text{g}) + E(\text{g}) &\rightarrow F(\text{liq})&\text{Byproduct,}\\
        3D(\text{g}) &\rightarrow 2F(\text{liq})&\text{Byproduct.}
    \end{aligned}\label{eq:tep_reactions}
\end{equation}
The \gls{tep} system is composed by five major process units: reactor, product condenser, vapor-liquid separator, recycle compressor and product stripper, shown in Fig.~\ref{fig:te_process_schematic}. Based on the reactions in equation~\ref{eq:tep_reactions}, there are 6 different \emph{modes of operation}, which correspond to 3 different $G/H$ mass ratios, and a desired product rate. The different modes of operation are shown in Table~\ref{tab:tep_operation_modes}.

From the perspective of \gls{da}, each mode of operation induces changes in the statistical properties of the data. As a result, a model learned with historical data from a set of operation modes (e.g., $1,\cdots,5$) may not generalize to a new operation mode (e.g., $6$). At the same time, collecting labeled data at the new operation mode is costly. \gls{msda} is thus a natural solution, where one leverages historical data from previous modes to learn a better model on the new mode, only requiring unlabeled data on the new operation conditions. In section~\ref{sec:preprocessing}, we describe a methodology for building a \gls{msda} benchmark on top of \gls{tep} simulations provided by~\cite{reinartz2021extended}.

\begin{table}[ht]
    \centering
    \caption{\gls{tep} operation modes, as described in~\cite{downs1993plant}. In our experiments, each mode of operation corresponds to a different domain.}
    \begin{tabular}{>{\centering}p{0.2\linewidth}>{\centering}p{0.2\linewidth}p{0.5\linewidth}}
        \toprule
         Mode & Mass Ratio & Production rate \\
         \midrule
         1 & 50/50 & 7038 kg h$^{-1}$ G and 7038 kg h$^{-1}$ H\\
         2 & 10/90 & 1408 kg h$^{-1}$ G and 12,669 kg h$^{-1}$ H\\
         3 & 90/10 & 10,000 kg h$^{-1}$ G and 1111 kg h$^{-1}$ H\\
         4 & 50/50 & maximum production rate\\
         5 & 10/90 & maximum production rate\\
         6 & 90/10 & maximum production rate\\
         \bottomrule
    \end{tabular}
    \label{tab:tep_operation_modes}
\end{table}

To build an \gls{afd} system, we need to collect data from a set of sensors, then categorize the data into a set of faults. In this paper, we use the data provided by~\cite{reinartz2021extended}. In their simulations, there are $53$ sensors in the overall plant, corresponding to different physical and chemical quantities. We group these variables into measurements (denoted XME$(i)$, for the $i-$th measurement) and manipulated (denoted XMV$(j)$, for the $j-$th manipulation), as shown in Table~\ref{tab:tep_variables}.

\begin{table}[ht]
    \centering
    \caption{Description of process variables of the \gls{tep}. Variables are divided into measurements (XME) and manipulated (XMV).}
    \resizebox{\linewidth}{!}{
\begin{tabular}{cccccccc}
    \toprule
    Variable         & Description                                        & Variable         & Description                                           & Variable         & Description                     & Variable         & Description                     \\ 
    \midrule
    XME(1)  & A Feed (kscmh)                                     & XME(15) & Stripper Level (\%)                                   & XME(29) & Component A in Purge (mol \%)   & XMV(2)  & E Feed (\%) \\
    XME(2)  & D Feed (kg/h)                                      & XME(16) & Stripper Pressure (kPa gauge)                         & XME(30) & Component B in Purge (mol \%)   & XMV(3)  & A Feed (\%) \\
    XME(3)  & E Feed (kg/h)                                      & XME(17) & Stripper Underflow (m$^{3}$/h)                        & XME(31) & Component C in Purge (mol \%)   & XMV(4)  & A \& C Feed (\%)        \\
    XME(4)  & A \& C Feed (kg/h)                                 & XME(18) & Stripper Temp (\textdegree C)          & XME(32) & Component D in Purge (mol \%)   & XMV(5)  & Compressor recycle valve (\%)   \\
    XME(5)  & Recycle Flow (kscmh)                               & XME(19) & Stripper Steam Flow (kg/h)                            & XME(33) & Component E in Purge (mol \%)   & XMV(6)  & Purge valve (\%)                \\
    XME(6)  & Reactor Feed rate (kscmh)                          & XME(20) & Compressor Work (kW)                                  & XME(34) & Component F in Purge (mol \%)   & XMV(7)  & Separator liquid flow (\%)      \\
    XME(7)  & Reactor Pressure (kscmh)                           & XME(21) & Reactor Coolant Temp (\textdegree C)   & XME(35) & Component G in Purge (mol \%)   & XMV(8)  & Stripper liquid flow (\%)       \\
    XME(8)  & Reactor Level (\%)                                 & XME(22) & Separator Coolant Temp (\textdegree C) & XME(36) & Component H in Purge (mol \%)   & XMV(9)  & Stripper steam valve (\%)       \\
    XME(9)  & Reactor Temperature (\textdegree C) & XME(23) & Component A to Reactor (mol \%)                       & XME(37) & Component D in Product (mol \%) & XMV(10) & Reactor coolant (\%)            \\
    XME(10) & Purge Rate (kscmh)                                 & XME(24) & Component B to Reactor (mol \%)                       & XME(38) & Component E in Product (mol \%) & XMV(11) & Condenser Coolant (\%)          \\
    XME(11) & Product Sep Temp (\textdegree C)    & XME(25) & Component C to Reactor (mol \%)                       & XME(39) & Component F in Product (mol \%) & XMV(12) & Agitator Speed (\%)             \\
    XME(12) & Product Sep Level (\%)                             & XME(26)          & Component D to Reactor (mol \%)                       & XME(40) & Component G in Product (mol \%)                     &         &                                 \\
    XME(13) & Product Sep Pressure (kPa gauge)                   & XME(27) & Component E to Reactor (mol \%)                       & XME(41) & Component H in Product (mol \%)                     &                  &                                 \\
    XME(14) & Product Sep Underflow (m$^{3}$/h)                  & XME(28) & Component F to Reactor (mol \%)                       & XMV(1)  & D Feed (\%)                     &                  &                           \\
    \bottomrule     
\end{tabular}
    }
    \label{tab:tep_variables}
\end{table}

In this dataset, the \gls{tep} system is simulated for a $100$ hours, with a sampling rate of $3$ minutes. As such, we use each simulation as a sample in our \gls{msda} benchmark. In each simulation, faults are introduced after $600$ time steps (i.e., $30$ hours). Concerning the type of faults, in their initial publication,~\cite{downs1993plant} presents 20 types of process disturbances (faults 1 through 20 in Table~\ref{tab:faults}), affecting different process variables. In addition to these initial faults,~\cite{bathelt2015revision} proposed 8 additional faults under the type \emph{random variation}, as shown in Table~\ref{tab:faults}.

\begin{table}[ht]
    \centering
    \caption{Description and types of faults for the \gls{tep} in the simulation environment of~\cite{reinartz2021extended}. Faults are grouped into $4$ types: step, random variation (RV), sticking and unknown.}
    \resizebox{\linewidth}{!}{
\begin{tabular}{cm{6cm}lcm{6cm}l}
\toprule
Fault & Variable                               & Type     & Fault Class & Variable                                                                                                                                               & Type     \\
\midrule
1           & A/C feed ratio, B composition constant & Step     & 15                              & Water outlet temperature (separator)                                                                                                                   & Sticking \\
2           & B composition, A/C ratio constant      & Step     & 16                              & Variation coefficient of the steam supply of the heat exchange of the stripper & RV       \\
3           & D feed temperature                     & Step     & 17                              & Variation coefficient of heat transfer (reactor)                                                                                                       & RV       \\
4           & Water inlet temperature (reactor)      & Step     & 18                              & Variation coefficient of heat transfer (condenser)                                                                                                     & RV       \\
5           & Water inlet temperature (condenser)    & Step     & 19                              & Unknown                                                                                                                                                & Unknown  \\
6           & A feed loss                            & Step     & 20                              & Unknown                                                                                                                                                & RV       \\
7           & C header pressure loss                 & Step     & 21                              & A feed temperature                                                                                                                                     & RV       \\
8           & A/B/C composition of stream 4          & RV       & 22                              & E feed temperature                                                                                                                                     & RV       \\
9           & D feed temperature 4                   & RV       & 23                              & A feed flow                                                                                                                                            & RV       \\
10          & C feed temperature                     & RV       & 24                              & D feed flow                                                                                                                                            & RV       \\
11          & Water outlet temperature (reactor)     & RV       & 25                              & E feed flow                                                                                                                                            & RV       \\
12          & Water outlet temperature (separator)   & RV       & 26                              & A \& C feed flow                                                                                                                                   & RV       \\
13          & Reaction kinetics                      & RV       & 27                              & Water flow (reactor)                                                                                                                                   & RV       \\
14          & Water outlet temperature (reactor)     & Sticking & 28                              & Water flow (condenser)                                                                                                                                 & RV     \\
\bottomrule
\end{tabular}}
    \label{tab:faults}
\end{table}

\subsection{Benchmark preparation}\label{sec:preprocessing}

\noindent\textbf{Data Cleaning.} For each mode, the simulations provided by~\cite{reinartz2021extended} are divided into $3$ groups: set-point variation, mode transitions and single fault. In the first case, the authors change the initial simulation set-point using a step or ramp function. In the second case, the simulation changes from one mode to another at an instant in time. In the third case, as previously mentioned, a fault is introduced at time step $600$, i.e., after 30 hours of simulation. For each fault, there are multiple intensities available (e.g., 25\%, 50\%, 75\% and 100\% fault magnitude). For magnitudes 25\%, 50\% and 75\%, the system is simulated 100 times, whereas for 100\%, the system is simulated 200 times. As a result, for the single-fault scenario only, the data provided by~\cite{reinartz2021extended} contains,
\begin{equation*}
    28\text{ faults}\times 6\text{ modes}\times 500\text{ simulations} = 84000\text{ simulations}.
\end{equation*}
Nonetheless, one should note that some simulations terminate earlier than $100h$, due to forced plant-shutdown. As a result, we adopt the following strategy: for each fault, we keep the first 100 simulations of highest magnitude that terminate successfully. For each selected simulation, we crop the signal into $2$ parts. The first 30h correspond to the steady state, determined by the set point of the mode of operation. This first part of the signal characterizes the healthy state of the system (i.e., faultless state). We further sub-sample the number of faultless state signals to keep a balanced dataset (i.e., 100 per mode of operation). The second part consists on the next 30h of simulation. Since faults are introduced exactly at the 601th time step, the second part of the signal characterize each fault. This process generates a slightly imbalanced dataset of 17289 samples\footnote{The data used in our experiments is available on Kaggle: \url{https://www.kaggle.com/datasets/eddardd/tennessee-eastman-process-domain-adaptation/data}}. We summarize the division of samples among modes of operation in Table~\ref{tab:samples}.

\begin{table}[ht]
    \centering
    \caption{Number and percentage of samples from each mode of operation.}
    \begin{tabular}{ccc}
        \toprule
        Mode of Operation & \# of Samples & \% of Samples \\
        \midrule
        1 & 2900 & 16.77\\
        2 & 2845 & 16.45\\
        3 & 2899 & 16.76\\
        4 & 2865 & 16.57\\
        5 & 2883 & 16.67\\
        6 & 2897 & 16.75\\
        Total & 17289 & 100\\
        \bottomrule
    \end{tabular}
    \label{tab:samples}
\end{table}
\noindent\textbf{Variable selection and pre-processing.} Out of the 53 variables presented in Table~\ref{tab:tep_variables} some of these variables are not continuous (e.g., XME$(23)$ through XME$(41)$). Given this remark, we follow~\cite{reinartz2021extended}, and consider a sub-set of 34 continuous signals as input to our neural nets. These are measurements XME$(1)$ through XME$(22)$, and manipulated variables XMV$(1)$ through XMV$(12)$. We thus have multi-variate time series of shape $(34, 600)$, where $34$ is the number of sensor readings (i.e., considered variables), and $600$ corresponds to the number of time steps ($T$). We further perform a standardization along each variable, within each mode, $x_{i,j,t}^{(P_{\ell})} = (x_{i,j,t}^{(P_{\ell})} - \mu_{j}^{(P_{\ell})})/\sigma_{j}^{(P_{\ell})}$, where,
\begin{align*}
    \mu_{j}^{(P_{\ell})} &= \dfrac{1}{n_{\ell}T}\sum_{i=1}^{n_{\ell}}\sum_{t=1}^{T}x_{i,j,t}^{(P_{\ell})}\text{, 
 and }\sigma_{i,j,t}^{(P_{\ell})} = \sqrt{\dfrac{1}{n_{\ell}T}\sum_{i=1}^{n_{\ell}}\sum_{t=1}^{T}(x_{i,j,t}^{(P_{\ell})}-\mu_{j}^{(P_{\ell})})^{2}},
\end{align*}
where $n_{\ell}$ is the number of samples for mode $\ell=1,\cdots,6$ (c.f., Table~\ref{tab:samples}).

\noindent\textbf{Neural Network Backbone.} In \gls{da}, it is common to choose a backbone upon which methods will rely on. For instance, in image processing, residual networks~\cite{he2016deep} are widely used. In the context of time series, In our paper, we employ a \gls{fcn}~\cite{long2015fully,wang2017time,ismail2019deep}, which consists on three convolutional blocks followed by a \gls{gap} layer. Each convolutional block has a convolutional layer, and a normalization layer. In our experiments we verified that instance normalization~\cite{ulyanov2016instance} improves stability and performance over other normalization layers such as batch normalization~\cite{ioffe2015batch}.
\subsection{Exploratory Data Analysis}


\begin{figure}[ht]
    \centering
    \begin{subfigure}{0.45\linewidth}
        \centering
        \includegraphics[width=\linewidth]{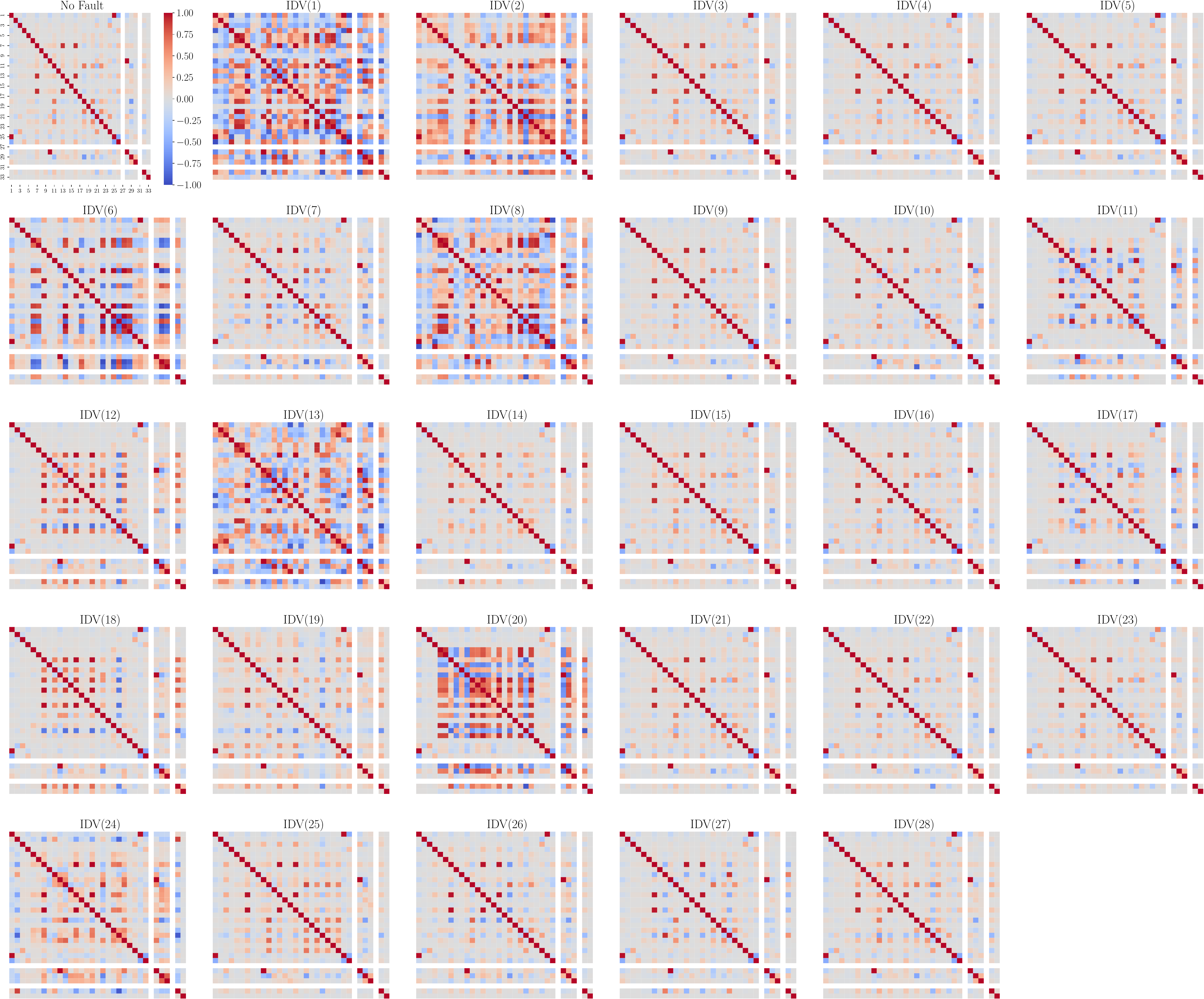}
        \caption{Mode 1}
    \end{subfigure}\hfill
    \begin{subfigure}{0.45\linewidth}
        \centering
        \includegraphics[width=\linewidth]{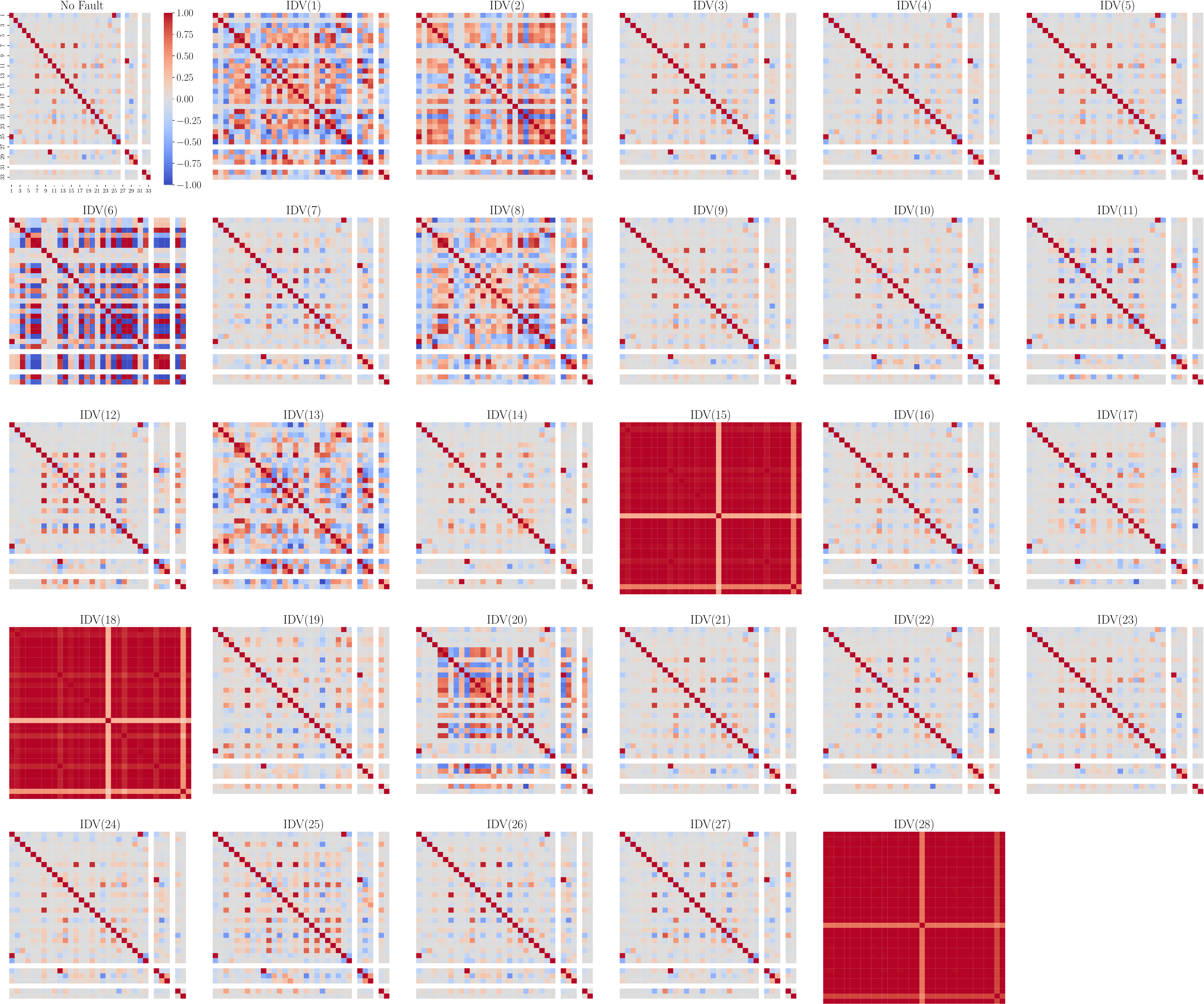}
        \caption{Mode 2}
    \end{subfigure}
    \caption{\textbf{Qualitative analysis of distributional shift.} In (a) and (b), we show the correlation between different variables in \gls{tep}, for modes 1 and 2, for each fault. On each correlation matrix, the coefficient $\rho_{jj'}$ corresponds to the Pearson correlation coefficient between $\{x_{j,t}\}_{t=1}^{600}$ and $\{x_{j',t}\}_{t=1}^{600}$ across simulations.}
    \label{fig:correlations}
\end{figure}

\noindent\textbf{Qualitative Analysis.} We analyze the pairwise correlations of variables, conditioned on the type of fault, that is, IDV$(1)$ through IDV$(28)$, and the no-fault scenario. In Fig.~\ref{fig:correlations}, we illustrate a change in the pattern of correlations between variables, conditioned on the fault type, for modes $1$ and $2$. In comparison, these patterns drastically change for faults 15, 18 and 28, corresponding to a sticking fault on the water outlet temperate on the separator, a random variation on the heat transfer coefficient on the condenser and a random variation on the water flow on the condenser, respectively. Hence, the mode of production deeply impacts the dynamic of the system, which creates a shift in distribution between data from these modes.

\begin{figure}[ht]
    \centering
    \begin{subfigure}{0.38\linewidth}
        \centering
        \includegraphics[width=\linewidth]{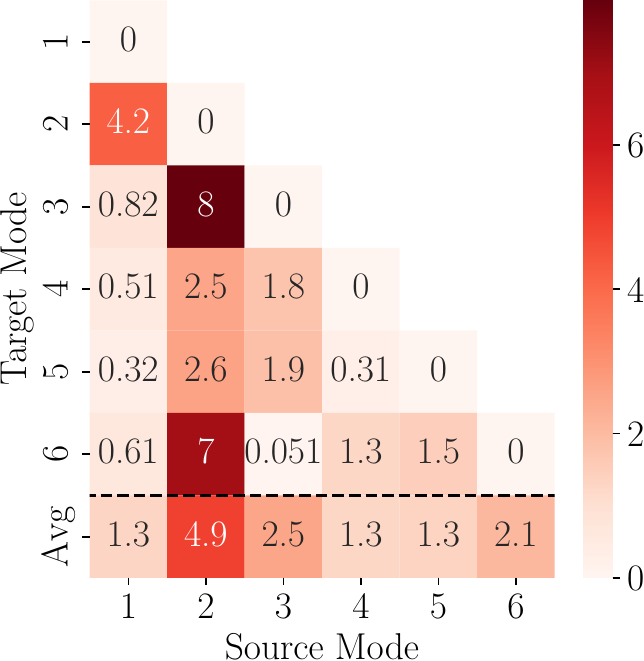}
        \caption{Pairwise $W_{2}$.}
    \end{subfigure}\hfill
    \begin{subfigure}{0.38\linewidth}
        \centering
        \includegraphics[width=\linewidth]{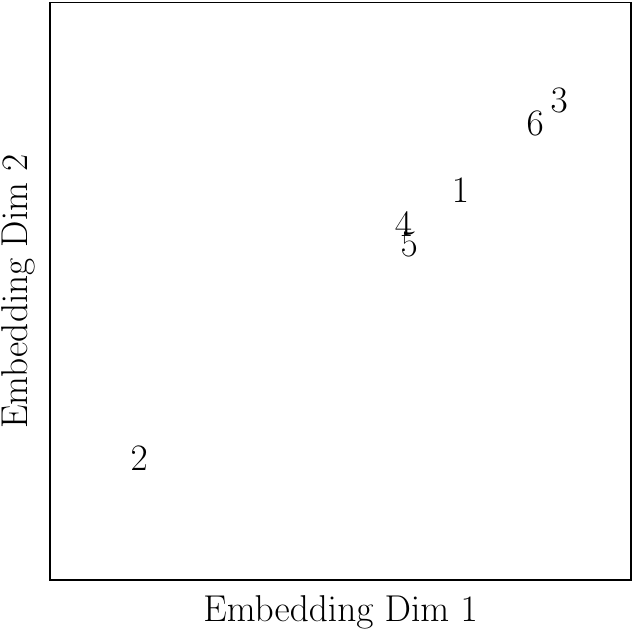}
        \caption{Mode embeddings.}
    \end{subfigure}
    \caption{\textbf{Quantitative analysis of distributional shift}. Pairwise Wasserstein distance between modes (a). Mode embeddings based on MDS (b).}
    \label{fig:distances}
\end{figure}

\noindent\textbf{Quantitative Analysis.} We quantify the shift between pairs of modes through the probability metrics introduced in section~\ref{sec:preliminaries}. We estimate the pairwise Wasserstein distances between modes using eq.~\ref{eq:w2_gauss}. In Fig.~\ref{fig:distances} (a), we show the pairwise distance in probability distribution between different modes. On one hand, the most different mode with respect others is Mode 2, which is especially far from modes $1$, $3$ and $5$. On the other hand, the most similar modes are $3$ and $6$. We can have a better picture about the level of similarity of these different domains by embedding them on the plane, as shown in Fig.~\ref{fig:distances} (b). We obtain these embeddings through \gls{mds}~\cite{kruskal1964multidimensional}, which defines the points in $\mathbb{R}^{2}$ while preserving the pairwise distances between the embeddings.

From our qualitative and quantitative analysis, we expect lower performances with respect the adaptation towards mode $2$, as it is the most dissimilar from other modes (c.f., Fig.~\ref{fig:distances} (a), average row). In contrast, adaptation between modes $(3, 6)$, and $(1, 4, 5)$ should work well as these modes share statistical characteristics. We verify these indications empirically in the next sections.

\subsection{Single-Source Domain Adaptation}

In this section, we explore single-source \gls{da}, i.e., when adaptation is done from a single source mode, to a single target mode. On the one hand, we refer to \emph{generalization}, to the ability of a classifier to perform well on unseen data from an unseen domain. On the other hand, we refer to \emph{adaptation}, when a classifier performs well on unseen data from the target domain.
\begin{figure}[ht]
    \begin{subfigure}{0.21\linewidth}
        \includegraphics[width=\linewidth]{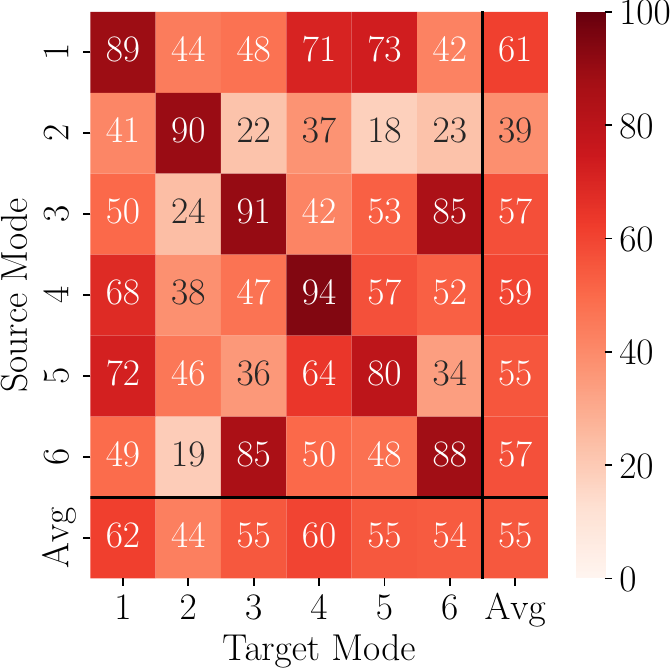}
        \caption{Baseline}
    \end{subfigure}\hspace{2mm}
    \begin{subfigure}{0.21\linewidth}
        \includegraphics[width=\linewidth]{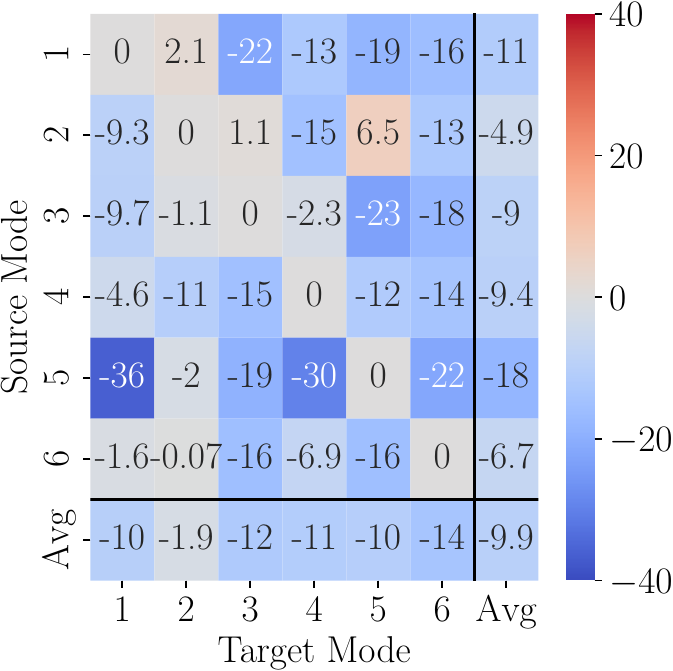}
        \caption{DANN}
    \end{subfigure}\hspace{2mm}
    \begin{subfigure}{0.21\linewidth}
        \includegraphics[width=\linewidth]{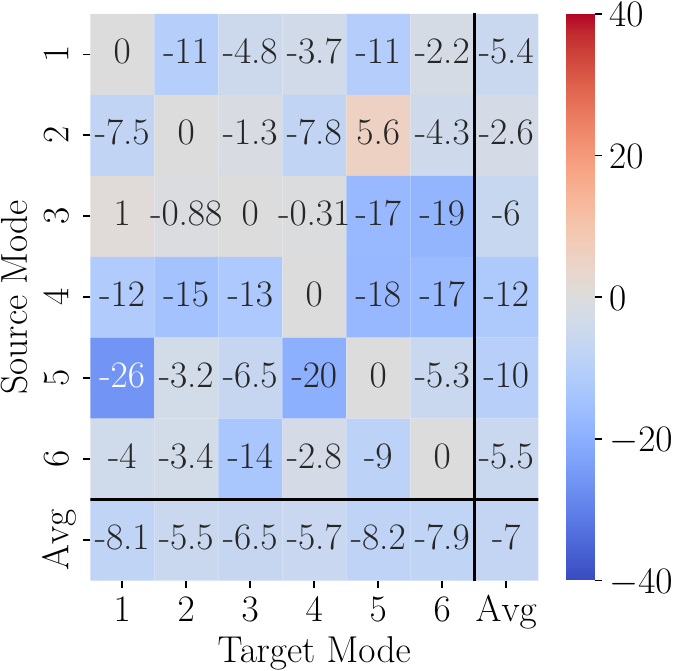}
        \caption{TCA}
    \end{subfigure}\hspace{2mm}
    \begin{subfigure}{0.21\linewidth}
        \includegraphics[width=\linewidth]{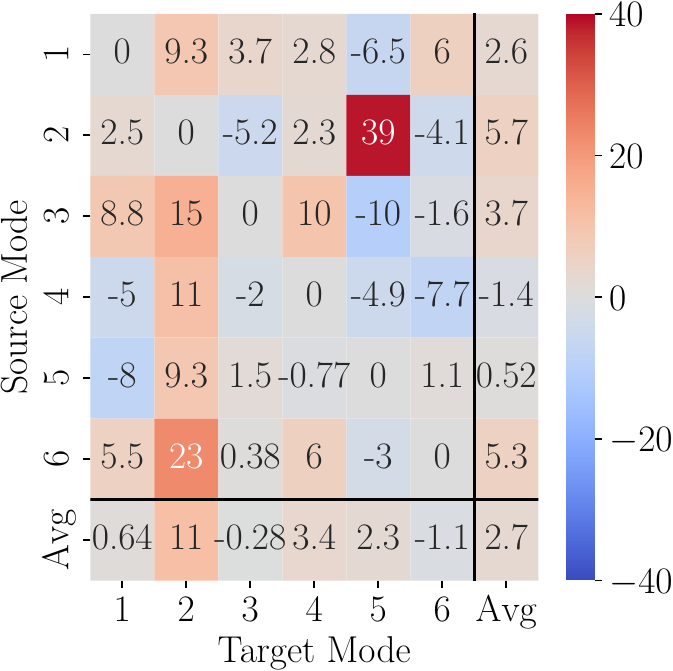}
        \caption{MMD}
    \end{subfigure}\\
    \begin{subfigure}{0.21\linewidth}
        \includegraphics[width=\linewidth]{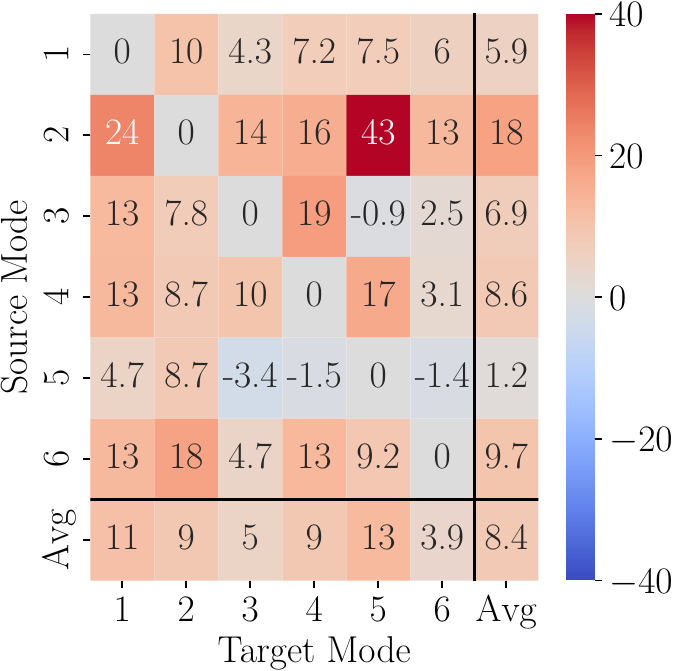}
        \caption{OTDA}
    \end{subfigure}\hspace{2mm}
    \begin{subfigure}{0.21\linewidth}
        \includegraphics[width=\linewidth]{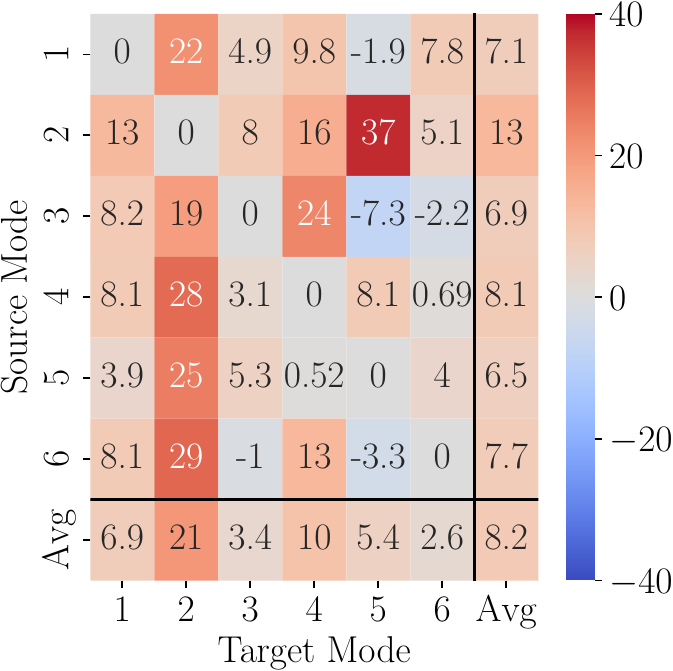}
        \caption{DeepJDOT}
    \end{subfigure}\hspace{2mm}
    \begin{subfigure}{0.21\linewidth}
        \includegraphics[width=\linewidth]{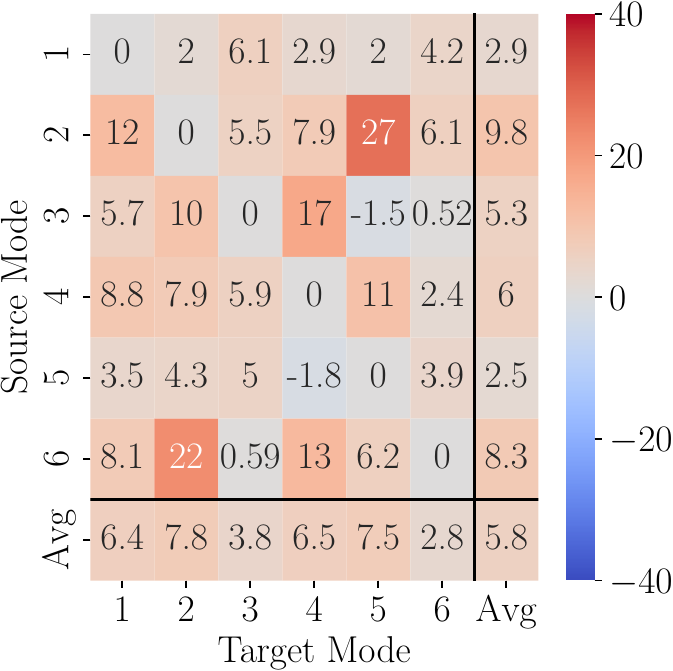}
        \caption{JDOT}
    \end{subfigure}
    \caption{Baseline (a) and single-source domain adaptation (b) algorithms.}
    \label{fig:overall_ssda}
\end{figure}
In this context, we have 2 baselines. The first, \emph{source-only}, considers that a classifier is learned exclusively with source domain data (i.e., no adaptation). This corresponds to the off-diagonals of Figure~\ref{fig:overall_ssda} (a). Second, we have \emph{target-only}, which trains and evaluates models on the target domain (i.e., no distributional shift). Note that the target-only scenario has an advantage over other methods, as it has access to labeled data in the target domain. This baseline can be seen as an upper bound in the adaptation performance.. With respect these scenarios, note that we verify our previous remarks, i.e., generalization towards mode $2$ is much more difficult than other domains, and the clusters of similar domains (e.g., $(3, 6)$ and $(1, 4, 5)$) generalize well.

We further compare the single-source \gls{da} methods presented in Table~\ref{tab:overview_da}, which are shown in Fig.~\ref{fig:overall_ssda} (b) through (g). Overall, we find that \gls{ot}-based methods have a higher performance than other metrics (e.g., the \gls{mmd}). This is similar to previous findings on smaller scale problems, such as~\cite{montesuma2022cross}. The best performing method is \gls{otda}~\cite{courty2016otda}, which maps source domain points to the target domain points through the \gls{ot} plan (c.f., eq.~\ref{eq:kantorovich}). Nonetheless, one should be mindful of \emph{negative transfer}~\cite{zhang2022survey} between similar modes (e.g., $3 \rightarrow 5$), which may result in performance degradation.

\subsection{Multi-Source Domain Adaptation}

\begin{figure}[ht]
    \centering
    \includegraphics[width=\linewidth]{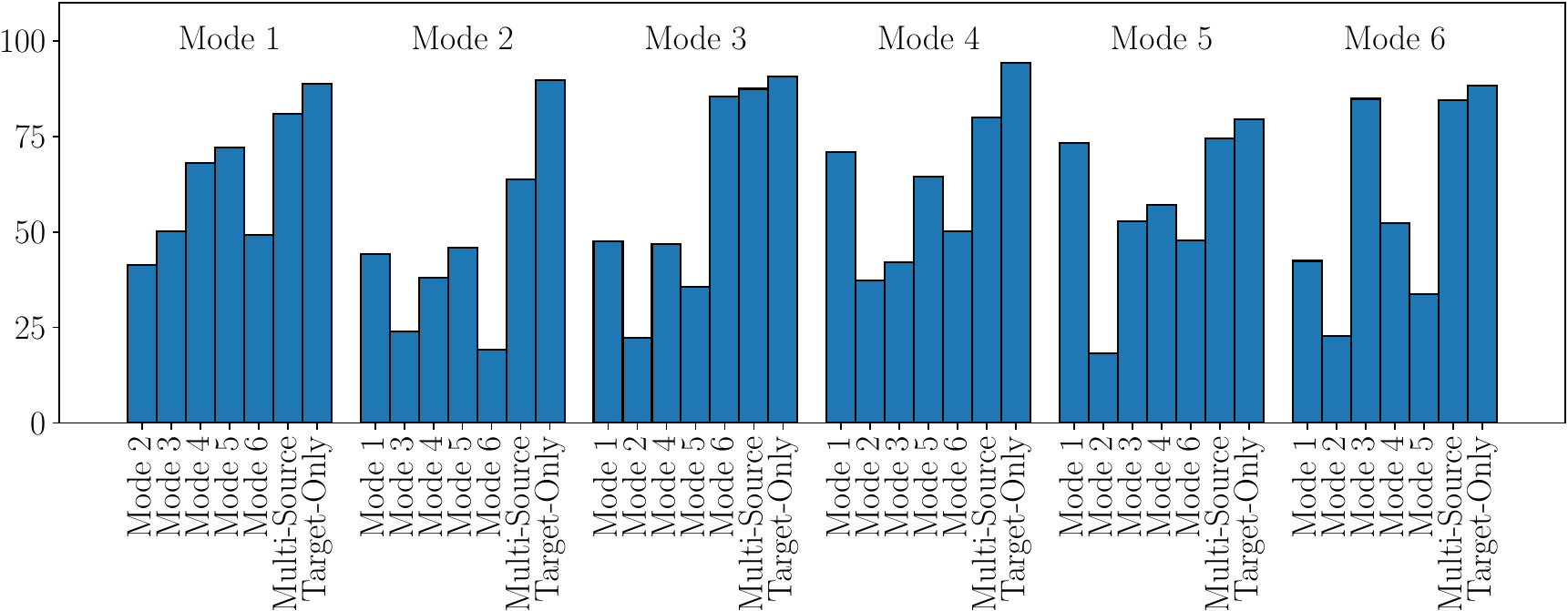}
    \caption{\textbf{Multi and single-source baseline comparison.} On top, we show the target domain. In the abscissa, we show the corresponding baseline. The multi-source scenario generally improves over the single source-only case.}
    \label{fig:baselines_comparison}
\end{figure}

In this section, we explore multi-source \gls{da}, i.e., when adaptation is done from multiple source domains towards a single target. Here, note that the models have much more labeled available data, as all domains are considered at once. We start our discussion by comparing the performance of single, source-only baselines, and the corresponding multi-source baseline for each target mode, which is shown in Fig.~\ref{fig:baselines_comparison}. Overall, the multi source-only baseline improve over single single-only for the same target. These baselines have similar performance when there are pairs of highly similar modes (e.g., modes $3$ and $6$), showing that extra data from additional modes is not as informative for generalization.

\begin{figure}[ht]
    \centering
    \includegraphics[width=\linewidth]{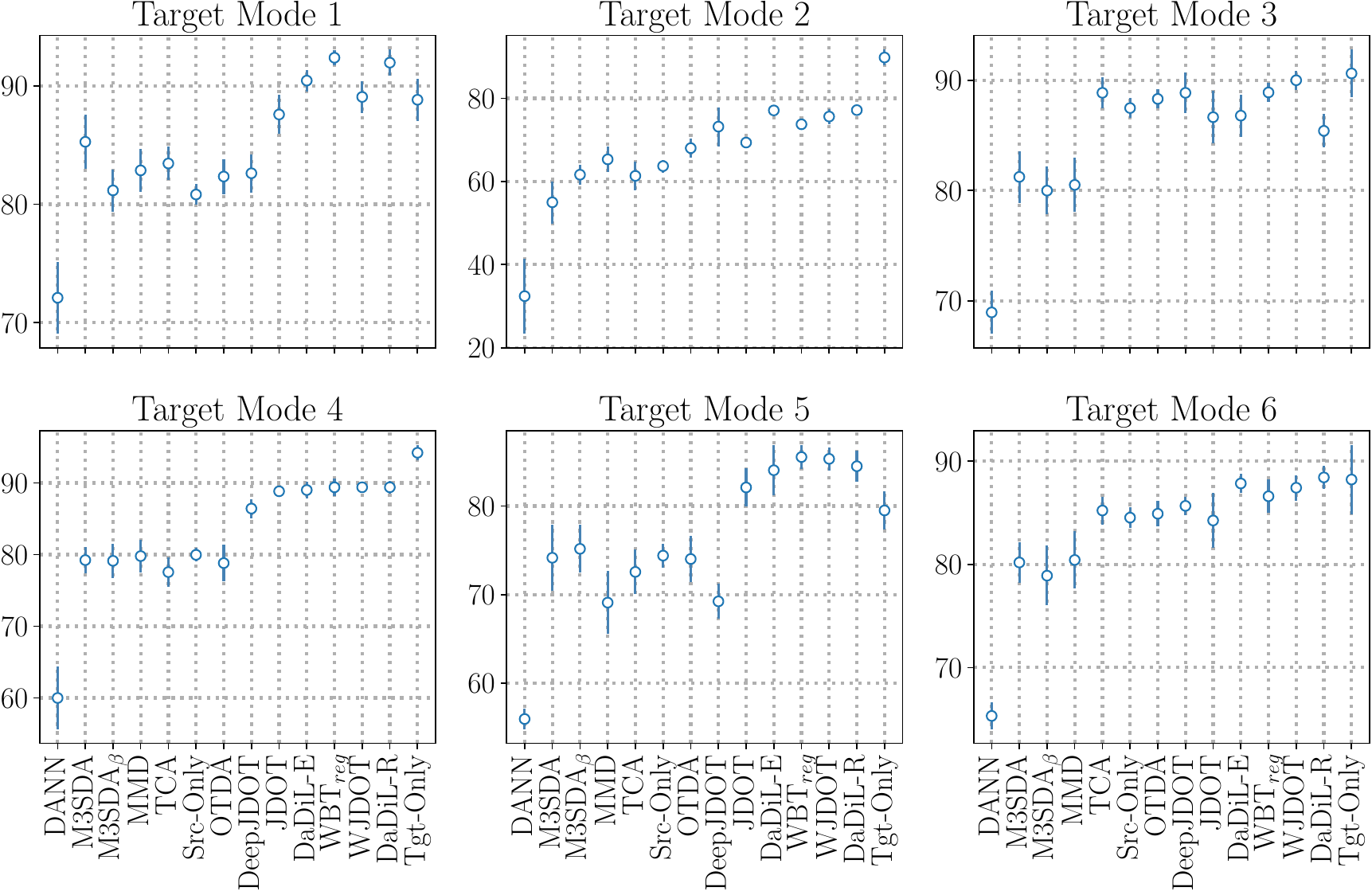}
    \caption{\textbf{Multi-source domain adaptation results.} We compare all algorithms with access to labeled data from all source domains, except the target mode, from which we have access to unlabeled data. Methods in the abcisssa are ordered by average performance on all modes.}
    \label{fig:tep_results}
\end{figure}

We now consider the performance of \gls{da} algorithms in the multi-source setting. Besides native \gls{msda} algorithms, i.e., algorithms that suppose the source as composed by different domains, we also consider single-source algorithms with access to the concatenation of all source domains. Our comparison is shown in Fig.~\ref{fig:tep_results}. A first question is whether access to additional data is beneficial to adaptation. For instance, in single-source \gls{da}, methods exhibited negative transfer in the task $3 \rightarrow 5$. When provided access to data from all domains, all single-source adaptation method performance improved over the single-source baseline. As a result, even though data from multiple domains may not improve generalization, it does improves adaptation.

With respect Fig.~\ref{fig:tep_results}, from the perspective of \gls{msda}, methods that weight sources in a linear space, such as WJDOT, or in a Wasserstein space, such as WBT and DaDiL outperform the weighting of classifiers' predictions, such as M3SDA$_{\beta}$~\cite{peng2019moment}. On the one hand, WJDOT can filter undesirable information during adaptation by assigning small weights to domains and samples. On the other hand, WBT and DaDiL combine the information in the sources non-linearly. These two strategies are effective in domain adaptation.

\begin{figure}[ht]
    \centering
    \includegraphics[width=0.8\linewidth]{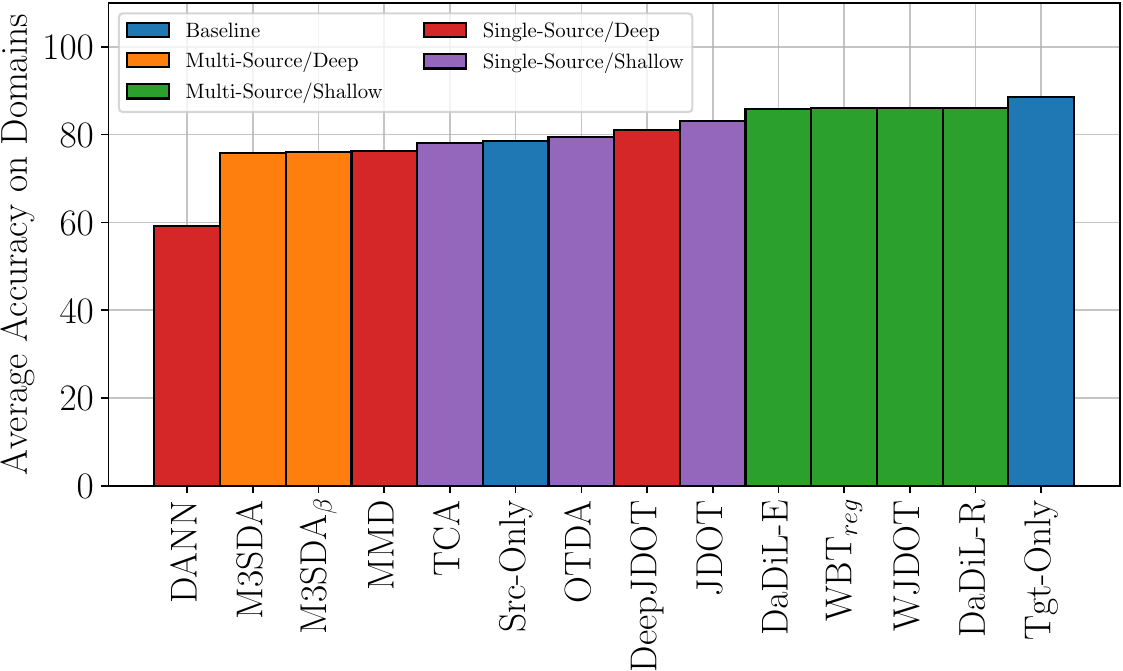}
    \caption{Comparison of average adaptation performance of \gls{da} algorithms in the multi-source setting.}
     \label{fig:avg_domain_perf}
\end{figure}

Finally, from Fig.~\ref{fig:avg_domain_perf}, we can see that shallow \gls{da} methods (e.g., JDOT) generally improve over deep \gls{da} methods (e.g., DeepJDOT). Indeed, deep \gls{da} methods learn features that are invariant to the domain shift between different modes. As a result, these features may be less useful for classification. In a general note (both single, and multi-source methods), \gls{ot}-based techniques outperform methods based on other distances, such as the MMD and $d_{\mathcal{H}}$. This remark agrees with previous studies on smaller scale systems~\cite{montesuma2022cross}.


\section{Conclusion}\label{sec:conclusion}

In this paper, we introduce a new benchmark for domain adaptation algorithms based on the Tenessee Eastman process~\cite{downs1993plant}. The present benchmark is created by applying pre-processing steps on the simulations provided by~\cite{reinartz2021extended} (c.f., section~\ref{sec:case_study}), thus creating a large scale dataset of time series. These time series are associated with different modes of production. Based on each mode of production, the statistical properties of the time series change (c.f., Fig.~\ref{fig:correlations}) creating a shift in the data probability distribution (c.f., Fig.~\ref{fig:distances}). As a result, data trained on a specific mode may not generalize well to other modes of production, thus the need for domain adaptation. Through a series of experiments with single-source and multi-source domain adaptation methods, we show that \gls{ot}-based methods outperform methods that rely on the maximum mean discrepancy, and $\mathcal{H}-$distances, which agrees with previous findings on smaller scale systems~\cite{montesuma2022cross}. Besides providing the open source code for the reproduction of our benchmark, with this work we hope to encourage research on the intersection between domain adaptation and fault diagnosis~\cite{zheng2019cross}.

%
\bibliographystyle{splncs04}
\bibliography{refs.bib}

\end{document}